%% file: main.tex
\documentclass{article}

\PassOptionsToPackage{numbers, sort, compress}{natbib}

\usepackage[preprint]{neurips_2025}

\usepackage[utf8]{inputenc} 
\usepackage[T1]{fontenc}    
\usepackage{hyperref}       
\usepackage{url}            
\usepackage{booktabs}       
\usepackage{amsfonts}       
\usepackage{nicefrac}       
\usepackage{microtype}      
\usepackage{xcolor}         
\usepackage{amsmath}
\usepackage{graphicx}
\usepackage{subcaption}

\usepackage{booktabs}
\usepackage{multirow}
\usepackage{array} 

\newcolumntype{C}[1]{>{\centering\arraybackslash}p{#1}}

\newcommand{\red}[1]{{\color{red}#1}}

\newcommand{\gray}[1]{{\color{gray}#1}}

\newcommand{\fref}[1]{Fig.~\ref{#1}}

\newcommand{\cref}[1]{Chapter~\ref{#1}}

\newcommand{\tref}[1]{Table~\ref{#1}}

\newcommand{\appref}[1]{Appendix~\ref{#1}}

\usepackage{pifont}
\newcommand{\cmark}{\ding{51}}%
\newcommand{\xmark}{\ding{55}}%

\title{C3R: Channel Conditioned Cell Representations for unified evaluation in microscopy imaging}

\author{
  Umar Marikkar\thanks{\url{u.marikkar@surrey.ac.uk}} \quad
  Syed Sameed Husain \quad
  Muhammad Awais \quad
  Sara Atito \\
  University of Surrey
}

\begin{document}

\maketitle

\begin{abstract}

Immunohistochemical (IHC) images reveal detailed information about structures and functions at the subcellular level. However, unlike natural images, IHC datasets pose challenges for deep learning models due to their inconsistencies in channel count and configuration, stemming from varying staining protocols across laboratories and studies. Existing approaches build channel-adaptive models, which unfortunately fail to support out-of-distribution (OOD) evaluation across IHC datasets and cannot be applied in a true zero-shot setting with mismatched channel counts. To address this, we introduce a structured view of cellular image channels by grouping them into either context or concept—where we treat the context channels as a reference to the concept channels in the image. We leverage this context-concept principle to develop Channel Conditioned Cell Representations (C3R), a framework designed for unified evaluation on in-distribution (ID) and OOD datasets. C3R is a two-fold framework comprising a channel-adaptive encoder architecture and a masked knowledge distillation training strategy, both built around the context-concept principle. We find that C3R outperforms existing benchmarks on both ID and OOD tasks, while a trivial implementation of our core idea also outperforms the channel-adaptive methods reported on the CHAMMI benchmark. Our method opens a new pathway for cross-dataset generalization between IHC datasets, without requiring dataset-specific adaptation or retraining.

\end{abstract}

\begin{figure}[t]
\centering
\includegraphics[width=0.80\linewidth]{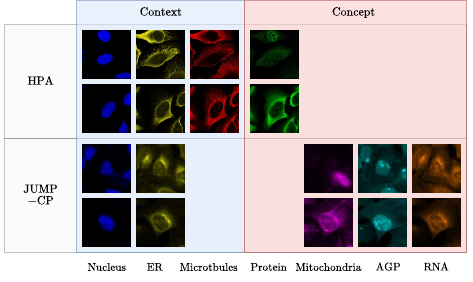}
\caption{The intrinsic separation of channels. Context channels serve as structural references and tend to exhibit high visual consistency across cells and datasets. Concept channels capture more variable, experiment-specific phenotypes, and exhibit greater diversity across instances.
}
\label{fig:dataset}
\end{figure}

\section{Introduction}

Immunohistochemical (IHC) imaging provides fine-grained visualizations of cells at the subcellular level, making it a valuable source of information used for biomedical applications such as disease prediction and drug discovery. Recently, representation learning using IHC images has gained prominence through self-supervised learning (SSL), and learned cell image representations have been evaluated on both in-distribution (ID) and out-of-distribution (OOD) tasks \citep{bray2016cell, caicedo2022cell, husain2023single, doron2023unbiased, gupta2024subcell, kim2025self}. However, unlike natural image datasets, IHC datasets differ in their immunofluorescent channel counts and configurations \citep{ILSVRC15, thul2017subcellular, chandrasekaran2023jump}. This poses a challenge for learning transferable representations, as typical vision encoders that build these representations operate under a fixed channel count. Existing studies address this either through model architecture or training strategy \citep{chen2023chammi, bao2023channel, pham2024enhancing}, but these methods are constrained by the channel configurations seen during training.

To address this, we first analyze the role of individual channels in IHC datasets and identify a natural division into two categories: context and concept \citep{thul2017subcellular}. Context channels offer structural reference information about the cell, such as nuclear shape or organelle organization. These channels are generally consistent and visually similar across images and datasets. In contrast, concept channels capture experiment-specific semantic information—like the subcellular location of a target protein in HPA \citep{thul2017subcellular}, or the response to a drug perturbation in JUMP-CP \citep{chandrasekaran2023jump}. These channels are the primary focus of downstream analysis. This distinction between context and concept has been previously explored in medical imaging- in radiology where the ribcage serves as structural context in X-rays, and in histopathology where hospitals provide specific stain contexts \citep{haghighi2022dira, bao2023contextual}. In contrast, \fref{fig:dataset} shows that for IHC images, this separation is even clearer, as the channels themselves can be grouped as context and concept.

On this insight, we introduce Channel Conditioned Cell Representations (C3R); to improve cell image representations and enable direct cross-dataset OOD evaluation. We propose a Context-Concept Encoder (CCE) architecture; that views cellular image channels as either context or concept, and processes these channels accordingly. The motivation of CCE stems from our hypothesis that the context group carries common information across datasets, hence learning the context separate from the concept yields transferable representations. We use these representations to perform zero-shot OOD evaluation without additional training. Moreover, we show that a naive implementation of our core idea- the architecture reduced to a single stem network- beats existing stem networks evaluated on the CHAMMI benchmark.

In addition to CCE, we propose Masked Context Distillation (MCD); a training strategy that encourages concept channels to infer from limited context. We design MCD based on the premise that the elements in context channels are spatially coherent. For example, \fref{fig:dataset} shows that the intensity of the E.R channel is typically more prominent on the boundary of the Nucleus. Hence, the concept channel should be able to contribute to a strong global representation, given either of the context channels.

The core contribution of C3R lies in its channel-adaptive encoder architecture and a novel context distillation strategy for SSL, outlined as follows:

\begin{itemize}
    \item \textbf{Context-Concept Encoder (CCE):} We introduce a channel-adaptive encoder, where based on the context-concept principle, a set group of channels are processed independently and then processed jointly. We use CCE for OOD evaluation across IHC datasets without requiring additional training, and we also show that the encoder outperforms existing ID and OOD baselines.
    \item \textbf{Masked Context Distillation (MCD):} Based on the same principle, we propose a self-supervised learning strategy for momentum-based encoders, where the student learns from a subset of context channels while the teacher accesses the full context. We find that the use of MCD leads to significant gains in ID performance.
\end{itemize}

\section{Related work}
\label{related_work}

\subsection{Representation learning for immunofluorescent images}

Recent advances in SSL for cell images span a range of approaches, from fully unsupervised methods to those incorporating weak or biological supervision. While some efforts focus on scaling models \citep{kenyon2024vitally,lorenci2025scaling,kraus2024masked}, we focus on architectural and SSL design choices. 

Several studies demonstrate the potential of SSL methods to learn representations from cell images without relying on external labels. \citet{kim2025self} train encoders using SimCLR, DINO, and MAE \citep{chen2020simple,caron2021emerging,he2022masked} on the JUMP-CP dataset \citep{chandrasekaran2023jump}, evaluating their learned features across diverse downstream tasks, where they find DINO to outperform other methods. Similar to \citet{kim2025self}, DINO4Cells \citep{doron2023unbiased} show that DINO learns unbiased cellular morphology features without domain-specific supervision, which is attributed to the zero-shot and linear probing capabilities inherent in clustering-based SSL methods like DINO and iBOT \citep{zhou2021ibot}. Here, the authors pre-train a separate model for each dataset to accommodate the different channel configurations. Other works \citep{xun2023microsnoop,murthy2024generalizable} explore alternative SSL strategies using reconstruction-based objectives.

Another subset of studies incorporate weak supervision to improve representations. CytoSelf \citep{kobayashi2022self} introduces a Protein ID-based loss within a VQ-VAE framework, while Set-DINO \citep{yao2024weakly} leverages structured labels from Optical Pooled Screening (OPS). BIDCell \citep{fu2024bidcell} employs RNA-seq data to supervise representations for segmentation tasks. SubCell \citep{gupta2024subcell} uses a supervised contrastive loss, treating antibody-stained cells as positive pairs. Similar to DINO4Cells, their model is trained on the HPA dataset \citep{thul2017subcellular}, and is evaluated on both HPA (ID) and JUMP-CP (OOD) datasets \citep{chandrasekaran2023jump}. To enable cross-dataset generalization, unlike DINO4Cells, Subcell pre-trains a new encoder on the same source dataset, with the channels adjusted to match the channel configuration of the target dataset.

Since DINO4Cells \citep{doron2023unbiased} and SubCell \citep{gupta2024subcell} are both trained and evaluated on HPA \citep{thul2017subcellular} and JUMP-CP \citep{chandrasekaran2023jump}, we select them as representative benchmarks for our evaluation. Both these methods re-train models to match OOD configurations. In contrast, we build models that are deployable on OOD datasets without training. Based on the existing studies \citep{kim2025self, doron2023unbiased}, we find that clustering-based SSL strategies \citep{caron2021emerging, zhou2021ibot} and domain-specific auxiliary losses \citep{gupta2024subcell} contribute most towards strong image representations. Hence, we choose to implement iBOT \citep{zhou2021ibot} along with the SubCell contrastive loss as our representation learning baseline.

\subsection{Vision transformers for multi-channel imaging}

Handling multi-channel data between datasets has been explored in the weather and climate domain, where each channel encodes distinct information such as temperature and precipitation \citep{9672063, nguyen2023climax}.

For multi-channel IHC datasets, the CHAMMI benchmark \citep{chen2023chammi} provides a framework for joint training and evaluation on different IHC datasets. It consists of over 220,000 single-cell images from three sources—each with different numbers of imaging channels, and nine downstream tasks on ID and OOD data. The benchmark is designed to compare between different stem networks that enable multi-channel training. Moving from stem networks to overall architectures, ChAdaViT \citep{bourriez2024chada} introduces a transformer architecture that tokenizes each input channel separately, using unique channel embeddings and shared positional embeddings. This method is extended by ChannelViT \citep{bao2023channel}, which proposes a hierarchical channel sampling strategy that selects a subset of channels without replacement. DiChaViT \citep{pham2024enhancing} builds on this by introducing a regularization loss that encourages sampling channels with minimally overlapping channel embeddings, promoting diversity between the sampled channels.

Though the above methods on IHC datasets support multi-channel, multi-dataset training, they are limited to channels seen during training, which stems from channel- or dataset-specific embedding initializations. Although these studies do propose adaptations to unseen channels with empty channel embeddings, they do not learn transferable relationships between channels. Additionally, ChannelViT and DiChaViT use channel sampling strategies where the required memory increases quadratically with the sampled channel count. Single-channel methods \citep{xun2023microsnoop, murthy2024generalizable, lian2025isolated, lorenci2025scaling} are proposed in subsequent studies, where each channel is passed independently in the forward pass during training. Although these models can be directly evaluated on unseen datasets without further training, they fail to capture inter-channel dependencies. Nonetheless, we include a single-channel approach in our evaluation for completeness.

\section{Methodology}
\label{method}

\subsection{C3R: Channel Conditioned Cell Representations}
\label{c3r}

We propose C3R, a framework based on the context–concept principle, which includes two main components: the Context-Concept Encoder (CCE) and the Masked Context Distillation (MCD) training strategy. The two components combined, encourages distinct feature learning for each context and concept, while encouraging the context to act as a reference for the concept. We implement C3R using a Vision Transformer (ViT) backbone and pre-train the model using iBOT \citep{dosovitskiy2020image, zhou2021ibot}.

First, we view a batch of $B$ input images $x$ as two groups of images, where $C_1, C_2$ are number of channels in the context and concept groups, and $(h,w)$ are the spatial dimensions. 
\begin{align}
\label{eq:1}
    x = \left[ x_{c1}, \, x_{c2} \right], \quad x_{c1} \in \mathbb{R}^{B \times C_1 \times h \times w}, \quad x_{c2} \in \mathbb{R}^{B \times  C_2 \times h \times w} .
\end{align}

\paragraph{Context-Concept Encoder (CCE).} 

We design CCE to construct separate intermediate representations for context and concept channels. Since context channels are typically consistent across datasets \citep{thul2017subcellular, chandrasekaran2023jump}, the intermediate context representations learned by the model transfer well to new datasets. As a result, in an out-of-distribution (OOD) setting, the context channels serve as a reliable reference for the concept channels in the target dataset.

\begin{figure}[t]
    \centering
    \begin{subfigure}[ht]{0.49\linewidth}
        \centering
        \includegraphics[height=5.6cm]{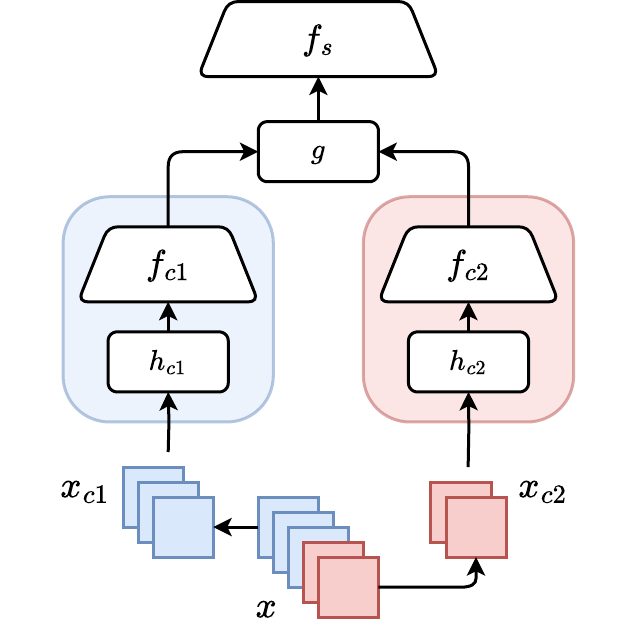}
        \caption{
        }
        \label{fig:arch}
    \end{subfigure}
    \hfill
    \begin{subfigure}[ht]{0.49\linewidth}
        \centering
        \includegraphics[height=5.6cm]{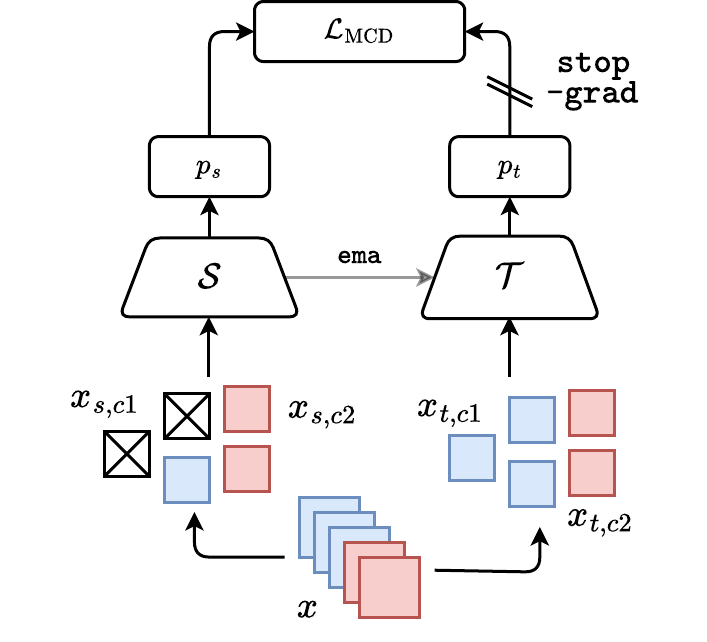}
        \caption{}
        \label{fig:mcd}
    \end{subfigure}
\caption{Overview of C3R. (a) Context-Concept Encoder: The input channels are separated into context and concept, where each group is processed independently through their respective \(h_c\) and \(f_c\) layers. The two group-wise representations are then combined and passed through a shared set of encoder layers \(f_s\). (b) Masked Context Distillation: During training, the student encoder \(\mathcal{S}\) randomly samples a subset of context channels prior to the forward pass, while the teacher encoder \(\mathcal{T}\) passes the full set of context channels. The loss is computed between the context-masked student representation and the dense teacher representation.}
    \label{fig:arch_mcd_combined}
\end{figure}

We define $h_{c1}$ and $h_{c2}$ to be context and concept convolutional stems, $f_{c1}$ and $f_{c2}$ to be context and concept encoder layers, and $f_s$ to be the shared encoder layers. \fref{fig:arch} shows the schematic of CCE.

Given the input \(x = \left[ x_{c1}, \, x_{c2} \right]\), we apply a channel-wise instance normalization to ensure that the information passed through each channel is not over or under-represented. Each channel in \( x_{c1}\) and \( x_{c2}\) is then tokenized by passing it through \(h_{c1}\) and \(h_{c2}\) respectively, yielding the feature maps,
\vspace{0.2cm}
\begin{align}
\tilde{x}_{c1} \in \mathbb{R}^{B \times C_1 \times N \times d}, \quad \tilde{x}_{c2} \in \mathbb{R}^{B \times C_2 \times N \times d} , 
\end{align}

where \( d \) is the output dimensionality of the \(h_{c1}\) and \(h_{c2}\), and \( N \) is the number of tokens after reshaping. \( d \) is set to \( D/2 \), where \( D \) is the embedding dimension of the baseline encoder. 

At this point, we introduce the branched encoders \(f_{c1}\) and \(f_{c2}\) for each group. As the individual channel feature maps need to be aggregated at some point to create a joint group feature map, we introduce a design choice regarding the stage at which the aggregation over channels is applied. Specifically, given the tokens \( \tilde{x}_{c1} \) and \( \tilde{x}_{c2} \), we consider two variants;

\begin{itemize}
    \item \textbf{Pre-aggregation}: We design \(f_{c1}\) and \(f_{c2}\) to encode joint group-wise feature maps, sharing information between channels within a group. We first apply mean pooling across channels to obtain a single group-wise feature map, then pass it through \(f_{c1}\) and \(f_{c2}\),
    \begin{align}
        \tilde{x}_{c1} = \texttt{MeanPool}(\tilde{x}_{c1}, \texttt{dim}=1), \quad 
        \tilde{x}_{c2} &= \texttt{MeanPool}(\tilde{x}_{c2}, \texttt{dim}=1) \\
        \hat{x}_{c1} = f_{c1}(\tilde{x}_{c1}), \quad 
        \hat{x}_{c2} &= f_{c2}(\tilde{x}_{c2}).
    \end{align}
    
    \item \textbf{Post-aggregation}: We design \(f_{c1}\) and \(f_{c2}\) to encode channel-wise feature maps, sharing information within each channel only. We apply \(f_{c1}\) and \(f_{c2}\) to each channel in \(\tilde{x}_{c1}\) and \(\tilde{x}_{c2}\) independently, then pool the outputs,
    \begin{align}
        \hat{x}_{c1} &= \texttt{MeanPool}\big(f_{c1}(\tilde{x}_{c1}), \texttt{dim}=1\big), \quad 
        \hat{x}_{c2} = \texttt{MeanPool}\big(f_{c2}(\tilde{x}_{c2}), \texttt{dim}=1\big).
    \end{align}
\end{itemize}

The output feature maps \(\hat{x}_{c1}\) and \(\hat{x}_{c2}\) obtained from \(f_{c1}\) and \(f_{c2}\) are then concatenated along the feature dimension \( d \). We then apply layer normalization, producing the final feature map \( \hat{x} \). This combined feature map is subsequently passed through the shared set of encoder layers \( f_s \) to obtain the final output representation $y$. 
\begin{align}
\hat{x} &= \mathrm{LN}\left( \texttt{Concat}\left( \hat{x}_{c1}, \, \hat{x}_{c2} \right) \right)  \in \mathbb{R}^{B \times N \times 2d}, \text{~where~} D = 2 \times d \\
y &= f_{s}\left( \hat{x} \right) .
\end{align}

The depth of \( f_s \) is chosen to match the total number of parameters of the baseline encoder for fair comparison. This representation \(y\) which is being generated by viewing the images as two groups which are common across datasets, is directly transferable to other datasets with different channel configurations.  

\paragraph{Masked context distillation (MCD).}
\label{sec:channel_mask}

MCD governs the interaction between context and concept channels during training, and builds a relationship between the individual context channels. Based on the structural coherence of context channels, Using MCD, we encourage the student concept channels to infer from limited context, building robust global representations. \fref{fig:mcd} shows the schematic of MCD. 

We define the student and teacher networks as \(\mathcal{S}\) and \(\mathcal{T}\), and each network contains a projection head defined as \(p_s\) and \(p_t\), respectively.

Given the input \(x = [x_{c1},\, x_{c2}]\), we randomly drop \(c\) channels from the context group \(x_{c1}\) during training, where \(1 \leq c < C_1\), to form the student and teacher inputs,
\begin{align}
x_{s,c1-\texttt{drop}} &= \texttt{DropChannels}(x_{c1}, c)  \in \mathbb{R}^{B \times (C_1-c) \times h \times w}, \quad \text{~where~} 0 \leq c < C_1 \\
x_s &= [x_{s,c1-\texttt{drop}},\, x_{s,c2}], \quad x_t = [x_{t,c1},\, x_{t,c2}].
\end{align}
Here, the teacher network always receives the full context without dropping. Furthermore, in addition to the channel dropping, we perform patch-wise random masking for all student channels as in iBOT \citep{zhou2021ibot}. \(x_s\) and \(x_t\) are then passed through their respective encoders \(\mathcal{S}\) and \(\mathcal{T}\) to obtain feature representations, and then passed through respective projection heads \(p_s\) and
\(p_t\) to obtain \(z_s\) and \(z_t\).
\begin{align}
    y_s &= \mathcal{S}(x_s), \quad y_t = \mathcal{T}(x_t), \\
    z_s &= p_s(y_s),  \quad z_t = p_t(y_t) .
\end{align}

The Masked Context Distillation loss then is computed as the KL-divergence between them.
\begin{align}
\mathcal{L}_{\text{MCD}} = \mathcal{L}_{\text{KL}}\big(z_s, z_t\big).
\end{align}

By separating and conditioning context and concept representations, C3R yields strong image representations and generalize to unseen IHC datasets, which we evaluate in the experiments below.

\subsection{Experiments}


\paragraph{Datasets and tasks.} The HPA dataset contains subcellular localization images of human proteins across various cell types, with annotated protein locations within individual cells. The total dataset contains 1,138,378 single-cell images, and were stained using a total of 11920 antibodies, which were used to split the dataset into training and validation. Each antibody stain highlights a unique combination of protein subcellular localizations in the image. HPA contains two protein localization tasks; a 19-class multi-label problem corresponding to classes in the Kaggle challenge \citep{hpa-single-cell-image-classification}, and a 31-class multi-label problem with a broader subset of labels including less prevalent classes.

The JUMP1 dataset contains 300 million cell images under both chemical and genetic perturbations \citep{chandrasekaran2024three}. We use JUMP-CP \citep{chandrasekaran2023jump}, a subset of JUMP1, filtered for chemical perturbations, resulting in 320 wells with chemical perturbation from one of 302 compounds, and 64 wells without any perturbation. The images subjected to a unique perturbation are regarded as the same class
for retrieval tasks.

The CHAMMI benchmark \citep{chen2023chammi} contains a total of 220k IHC images from WTC-11 \citep{viana2023integrated}, HPA \citep{thul2017subcellular} and JUMP-CP \citep{chandrasekaran2023jump}, with different channel configurations. The original benchmark evaluates channel-adaptive stem networks, and includes nine tasks across three datasets, with three ID tasks and six OOD tasks.

\paragraph{Model training evaluation.} For all experiments, we pre-train Vision Transformer (ViT) backbones on the HPA dataset \citep{thul2017subcellular} using iBOT \citep{zhou2021ibot} with the Subcell antibody loss \citep{gupta2024subcell}. As our baseline, we train a 4-channel model for HPA evaluation and a 3-channel model to evaluate on JUMP-CP. The 3-channel model is trained following SubCell \citep{gupta2024subcell}, where the Microtubules channel is dropped from the HPA dataset, and the Protein channel is shared between Mitochondria, AGP, and RNA in JUMP-CP during evaluation. The C3R encoder trained on the 4-channel HPA dataset is used for all evaluations, and no re-training or adaptation is carried out. The context and concept groups assigned for C3R for each dataset are chosen as in \fref{fig:dataset}.

For DINO4Cells and SubCell, we use the official checkpoints. For ChannelViT and DiChaViT, we adapt the tokenizer and loss function code from their repositories and pre-train their methods under our baseline setting. To evaluate ChannelViT and DiChaViT on JUMP-CP, we share the learned Protein channel embedding across the three channels in JUMP-CP, following the same intuition as SubCell \citep{gupta2024subcell}. To compare with single-channel methods \citep{lian2025isolated, lorenci2025scaling}, we implement SingleChan, where a single channel is randomly sampled per training step. At inference time, we mean-pool the independently processed CLS tokens from all channels.

For downstream tasks, we follow SubCell, where we train a 3-layer MLP in a multi-label setting for HPA protein localization, and evaluate JUMP-CP performance using zero-shot retrieval and kNN evaluation on drug perturbation matching. We use the official CHAMMI \citep{chen2023chammi} repository to insert our grouped stem network (the combination of \(h_{c1}\) and \(h_{c2}\)) onto a ConvNeXt \citep{liu2022convnet} encoder pre-trained on ImageNet-22k \citep{ILSVRC15}. We directly concatenate the stem network outputs along the feature dimension before passing it to the ConvNext backbone. For comparison, we re-implement four top-performing reported variants in CHAMMI—SliceParam, TargetParam, TemplateMixing \citep{savarese2019learning}, and HyperNet \citep{ha2016hypernetworks}—using their recommended hyper-parameters and the same backbone.

\paragraph{C3R implementation details.} The C3R encoder CCE, uses the same ViT blocks as a vanilla ViT, with the added branching mechanism. Hence, we refer to the C3R backbones also as ViTs. For both backbones (ViT-S and ViT-B), we set two layers per branch, and 10 shared encoder layers. We use the post-aggregation variant for all experiments when using combined C3R (CCE + MCD), and the pre-aggregation variant for
experiments only CCE.

Further implementation details, and the choice of layer depths and aggregation variants are explained in the supplementary material.

\section{Results}

\begin{table}[t]
\centering
\caption{Performance comparison on HPA and JUMP-CP benchmarks. \underline{underline}: re-trained from scratch to match the JUMP-CP channel configuration.}
\label{tab:hpa_jumpcp_results}
\begin{tabular}{llccccc}
\toprule
\multirow{2}{*}{Encoder} & \multirow{2}{*}{Method} & \multicolumn{2}{c}{HPA-mAP (ID)}  & \multicolumn{2}{c}{JUMP-CP (OOD)} & \multirow{2}{*}{Average}  \\
\cmidrule(lr){3-4} \cmidrule(lr){5-6}
&& 31-loc & 19-loc & mAP & kNN &  \\
\midrule
\multirow{6}{*}{ViT-S}

&Baseline      
& 0.505 {\scriptsize$\pm$0.002} & 0.686 {\scriptsize$\pm$0.002} 
& \underline{0.355} & \underline{0.507} 
& 0.513 \\

&SingleChan         
& 0.380 {\scriptsize$\pm$0.002} & 0.528 {\scriptsize$\pm$0.002} & 0.327 & 0.457 & 0.423 \\

&ChannelViT         
& 0.438 {\scriptsize$\pm$0.001} & 0.602 {\scriptsize$\pm$0.001} 
& 0.345 & 0.503 
& 0.472 \\

&DiChaViT           
& 0.429 {\scriptsize$\pm$0.002} & 0.590 {\scriptsize$\pm$0.001} 
& 0.343 & 0.494 
& 0.464 \\

\cmidrule(lr){2-7}
&CCE 
& 0.520 {\scriptsize$\pm$0.002} & 0.705 {\scriptsize$\pm$0.003} 
& \textbf{0.358} & \textbf{0.530} 
& 0.528 \\

&C3R                
& \textbf{0.536} {\scriptsize$\pm$0.004} & \textbf{0.722} {\scriptsize$\pm$0.004} 
& 0.354 & 0.518 
& \textbf{0.533} \\

\midrule
\multirow{6}{*}{ViT-B}

&Baseline     
& 0.515 {\scriptsize$\pm$0.001} & 0.698 {\scriptsize$\pm$0.002} 
& \underline{0.355} & \underline{0.513} 
& 0.520 \\

&SingleChan         
& 0.385 {\scriptsize$\pm$0.002} & 0.528 {\scriptsize$\pm$0.001} 
& 0.339 & 0.473 
& 0.432 \\

&DINO4Cells         
& 0.508 {\scriptsize$\pm$0.000} & 0.683 {\scriptsize$\pm$0.000} 
& \underline{0.339} & \underline{0.509} 
& 0.510 \\

&SubCell            
& 0.519 {\scriptsize$\pm$0.002} & 0.695 {\scriptsize$\pm$0.002} 
& \underline{0.350} & \underline{0.514} 
& 0.520 \\

\cmidrule(lr){2-7}
&CCE       
& 0.531 {\scriptsize$\pm$0.002} & 0.716 {\scriptsize$\pm$0.002} 
& 0.358 & \textbf{0.532} 
& 0.534 \\

&C3R                
& \textbf{0.548} {\scriptsize$\pm$0.002} & \textbf{0.737} {\scriptsize$\pm$0.003} 
& \textbf{0.363} & 0.530 
& \textbf{0.544} \\

\bottomrule
\end{tabular}

\end{table}

\subsection{Benchmarks}

\paragraph{Representation learning benchmarks.} \tref{tab:hpa_jumpcp_results} compares the performance of existing methods and C3R on the HPA and JUMP-CP benchmarks. From \tref{tab:hpa_jumpcp_results}, we observe that for both ViT-S and ViT-B backbones, C3R achieves the highest average performance outperforming all other baselines. We find that pretraining CCE under the baseline setting without MCD also outperforms existing benchmarks, highlighting the effectiveness of the architecture itself. Although ChannelViT and DiChaViT improve on the naive single-channel training strategy, it fails to match the iBOT baseline. We attribute this to two features that are common in these methods; the shared stem networks over all channels not being able to learn channel-specific information from the beginning, and the concept channels being randomly sampled out during training, which in essence, results in non-informative training samples.


\paragraph{CHAMMI benchmark for channel-adaptive models.} We apply a naive implementation of our core idea- The grouped embedding stem in C3R without branches or MCD- on the CHAMMI benchmark. \tref{tab:chammi} compares the performance of existing stem networks and our method. We find that our method ranks highest in 4 out of the 9 individual tasks, while exhibiting the highest metrics in ID and OOD averages, and on the CHAMMI Performance Score (CPS). Although the main objective of using C3R is to learn strong representations that support unified evaluation, we show that our naive implementation still outperforms the reported methods in CHAMMI.

\begin{table}[t]
\centering
\caption{Comparisons of stem networks on the CHAMMI benchmark. SP, TP, TM and HN stand for SliceParam, TargetParam, TemplateMixing and HyperNet, respectively.}
\label{tab:chammi}
\begin{tabular}{ll*{5}{c}}
\toprule
\multirow{2}{*}{Dataset} & \multirow{2}{*}{Task} & \multicolumn{5}{c}{F1 Score} \\
\cmidrule(lr){3-7}
& & SP & TP & TM & HN & ours \\ 
\midrule
\multirow{2}{*}{WTC-11} 
    & Task 1 (ID)  & 0.827 & 0.834 & 0.850 & 0.864 & \textbf{0.874} \\
    & Task 2 (OOD) & 0.760 & 0.813 & 0.827 & \textbf{0.847} & 0.838 \\
\midrule
\multirow{3}{*}{HPA} 
    & Task 1 (ID)  & 0.920 & 0.920 & \textbf{0.930} & 0.928 & 0.919 \\
    & Task 2 (OOD) & 0.897 & 0.900 & \textbf{0.907} & 0.903 & 0.895 \\
    & Task 3 (OOD) & 0.587 & 0.564 & 0.586 & 0.592 & \textbf{0.625} \\
\midrule
\multirow{4}{*}{JUMP-CP} 
    & Task 1 (ID)  & 0.852 & 0.824 & 0.852 & 0.860 & \textbf{0.888} \\
    & Task 2 (OOD) & \textbf{0.579} & 0.567 & 0.547 & 0.573 & 0.524 \\
    & Task 3 (OOD) & \textbf{0.232} & 0.221 & 0.209 & 0.179 & 0.218 \\
    & Task 4 (OOD) & 0.082 & 0.082 & 0.078 & 0.075 & \textbf{0.096} \\
\midrule
\multicolumn{2}{l}{Average ID}  & 0.866 & 0.859 & 0.878 & 0.884 & \textbf{0.893} \\
\multicolumn{2}{l}{Average OOD} & 0.523 & 0.525 & 0.526 & 0.528 & \textbf{0.533} \\
\multicolumn{2}{l}{CHAMMI score (CPS)} & 0.600 & 0.612 & 0.618 & 0.623 & \textbf{0.626} \\
\bottomrule
\end{tabular}

\end{table}

\subsection{Analysis}

\paragraph{The contribution of each component of C3R.}
\label{sec:C3Rablation}

\tref{tab:component_analysis} shows the incremental contribution of each component in C3R over the baseline ViT. We first observe that the grouped convolutional stem network (GC) alone improves ID performance on HPA. However, when directly applied to JUMP-CP, it is unable to match the baseline OOD result. Adding the branched network, which processes input groups independently for a few layers, yields modest ID gains but more importantly, significantly boosts OOD performance, matching or exceeding the JUMP-CP baseline.

Introducing instance normalization before convolutions (IN) further enhances OOD generalization, improving over the baseline. This suggests that refining group-wise low-level representations through branching should be carried out while preventing individual channels from being over- or under-emphasized through instance norms. The above elements create the overall architecture of CCE. Using masked context distillation (MCD), where some context channels are dropped during training and the student network benefits from the teacher's full context channel representation, yields significant improvements in ID performance. However, the ID performance boost obtained by MCD does not translate to the OOD task. This will be investigated in future work.

\begin{table}[t]
\centering
\caption{The effect of each component in C3R. The \underline{underlined} methods have been pre-trained to match the JUMP-CP channel configuration. 'GC': grouped stem, 'IN': instance norm, 'B': branched architecture, 'MCD': context distillation. GC, IN and B combine to create the CCE architecture.}
\label{tab:component_analysis}
\begin{tabular}{l l c c l c c l}
\toprule

\multirow{2}{*}{Enc.} & \multirow{2}{*}{Method} & \multicolumn{3}{c}{HPA-mAP (ID)}  & \multicolumn{3}{c}{JUMP-CP (OOD)}  \\
\cmidrule(lr){3-5} \cmidrule(lr){6-8}
& & 31-loc & 19-loc & \multicolumn{1}{c}{Avg.} & mAP & kNN & \multicolumn{1}{c}{Avg.} \\
\midrule
\multirow{5}{*}{ViT-S} 
& Baseline         & 0.505 & 0.686 & 0.596 & \underline{0.355} & \underline{0.507} & \underline{0.431} \\
& +GC              & 0.523 & 0.700 & 0.612 \small\textcolor{green!60!black}{↑ 0.016} & 0.347 & 0.510 & 0.429 \small\textcolor{red!70!black}{↓ 0.002} \\
& +GC+B            & 0.521 & 0.704 & 0.613 \small\textcolor{green!60!black}{↑ 0.017} & 0.350 & 0.521  & 0.435 \small\textcolor{green!60!black}{↑ 0.004} \\
& +GC+IN+B           & 0.520 & 0.705 & 0.613 \small\textcolor{green!60!black}{↑ 0.017} & \textbf{0.358} & \textbf{0.530} & \textbf{0.444} \small\textcolor{green!60!black}{↑ 0.013} \\
& +GC+IN+B+MCD        & \textbf{0.535} & \textbf{0.725} & \textbf{0.630} \small\textcolor{green!60!black}{↑ 0.034}  & 0.354 & 0.518 & 0.436 \small\textcolor{green!60!black}{↑ 0.005} \\

\midrule
\multirow{5}{*}{ViT-B} 
& Baseline         & 0.515 & 0.698 & 0.607 & \underline{0.355} & \underline{0.513} & \underline{0.434} \\
& +GC              & 0.529 & 0.710 & 0.620 \small\textcolor{green!60!black}{↑ 0.013} & 0.344 & 0.508 & 0.426 \small\textcolor{red!70!black}{↓ 0.008} \\
& +GC+B            & 0.530 & 0.721 & 0.626 \small\textcolor{green!60!black}{↑ 0.019} & 0.352 & 0.513 &  0.432 \small\textcolor{red!70!black}{↓ 0.002} \\
& +GC+IN+B               & 0.531 & 0.716 & 0.623 \small\textcolor{green!60!black}{↑ 0.017} & 0.358 & \textbf{0.532} & 0.445 \small\textcolor{green!60!black}{↑ 0.011} \\
& +GC+IN+B+MCD        & \textbf{0.548} & \textbf{0.737} & \textbf{0.642} \small\textcolor{green!60!black}{↑ 0.036} & \textbf{0.363} & 0.530 & \textbf{0.446} \small\textcolor{green!60!black}{↑ 0.012} \\

\bottomrule
\end{tabular}

\end{table}

\begin{figure}[t]
\centering
\begin{minipage}[]{0.4\linewidth}
    \centering
    \includegraphics[width=\linewidth]{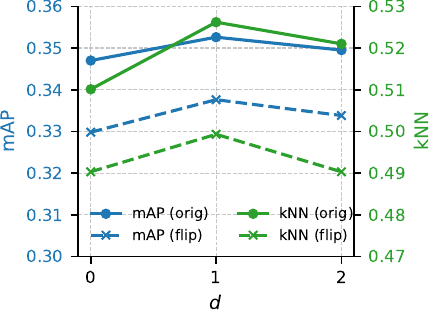}
    \caption{Effects of group switching assignments. The experiments were carried out for JUMP-CP using ViT-S without MCD and \(d\) layers per branch.}
    \label{fig:jump_flip}
\end{minipage}
\hfill
\begin{minipage}[]{0.55\linewidth}
    \centering
    \captionof{table}{Choice of channel dropping for MCD. The experiments were carried out using ViT-S and the post-aggregation variant. None: no dropping, S,T: dropping channels in student and teacher, respectively.}
    \label{tab:mcd}
    \begin{tabular}{l c c c c}
    \toprule
    \multirow{2}{*}{Drop} & \multicolumn{2}{c}{HPA-mAP} & \multicolumn{2}{c}{JUMP-CP} \\
    \cmidrule(lr){2-3} \cmidrule(lr){4-5}
     & 31-loc & 19-loc & mAP & kNN \\
    \midrule
    SingleChan     &   0.380 &  0.528    &   0.327    &    0.457    \\
    ChannelViT     & 0.438     &   0.602     &  0.345     & 0.503       \\
    DiChaViT       & 0.429     &   0.590     &   0.343    & 0.494       \\
    \midrule
    None           & 0.519 & 0.702 & 0.351 & \textbf{0.530} \\
    S              & \textbf{0.536} & 0.722 & \textbf{0.354} & 0.518 \\
    S+T            & 0.533 & \textbf{0.724} & 0.347 & 0.504 \\
    \bottomrule
    \end{tabular}
\end{minipage}
\end{figure}

\paragraph{The foundational assumption of group-wise learning and MCD.}
\label{sec:mcd_results}

To validate our hypothesis that the context and concept groups, and their respective branched layers encode distinct information, we run OOD experiments by deliberately switching group assignments during the forward pass of the model. A significant drop in performance under this perturbation would suggest that the branched layers are group-specific, thereby supporting our hypothesis. Conversely, if the performance remains similar, it would imply that the branches are group-agnostic and do not encode distinct information. The observations in \fref{fig:jump_flip} appear to be in line with our hypothesis, where we observe a consistent drop in performance when the branches are flipped.

Our hypothesis for MCD lies in encouraging the concept channels of the model to contribute to the overall image representation with limited context. To validate this hypothesis, we run experiments to identify the networks (student or teacher) on which the channels need to be dropped in order to yield better representations. \tref{tab:mcd} compares the performance of dropping context channels under different settings. We also compare the results versus SingleChan, ChannelViT and DiChaViT, which we categorize as context-concept agnostic channel dropping strategies.

From \tref{tab:mcd}, we observe that ensuring the concept channels are preserved during the forward pass results in better performance (None, S and S+T), compared to the context-concept agnostic sampling strategies. We also find that over no sampling at all, sampling the context under either setting (S or S+T) yields better ID performance. This observation aligns with our motivation of using MCD, where a masked context generally outperforms a non-masked context channel set. However, we observe a drop in average performance when the teacher context channels are masked (S+T). We attribute this to the limited learning capability caused by the weak teacher signal, as seen in existing SSL methods where masked teachers \citep{chen2022sdae} under perform in linear evaluation tasks in comparison to methods with full teacher representations
\citep{caron2021emerging, zhou2021ibot}.

\section{Broader Impacts and limitations}

\paragraph{Broader Impact.}

IHC imaging plays a critical role in clinical diagnostics and biomedical research, providing key insights into cellular morphology, protein localization, and disease progression at the subcellular level. However, the integration of deep learning methods towards these diagnostics has been limited by the lack of generalizable models that can operate reliably across heterogeneous datasets, where retraining models for every lab, institution, or imaging protocol is impractical and costly. Without dataset-specific adaptation or re-training, C3R has the potential to accelerate the integration of deep learning into diagnostic pipelines, reducing the time and cost of biomarker discovery, drug response prediction, and personalized treatment planning. However, as with all AI systems applied to biomedical data, misuse is possible. Malicious actors could potentially exploit representations from our pre-trained models for unauthorized diagnosis. This highlights the need of using models like C3R responsibly, with clear validation and compliance with ethical guidelines.

\paragraph{Limitations.} Based on our primary context-concept assumption, the ability to perform OOD evaluation on a target dataset depends on the assumption being valid for that dataset i.e: we find a natural separation of channels into context and concept in the target dataset. In the most common and publicly available IHC datasets (HPA \citep{thul2017subcellular}, JUMP \citep{chandrasekaran2023jump}, WTC-11 \citep{viana2023integrated}, OpenCell \citep{cho2022opencell}, Bridge2AI \citep{clark2024cell}) we find this assumption to be true. However, we have not explored IHC datasets that may not follow this assumption.

The CCE architecture serves as a proof of concept for our core assumption, and more elegant ways of building up the concept representation based on the context (e.g.: cross-attention \citep{bao2023contextual}) need to be explored. Furthermore, like ChannelViT and DiChaViT \citep{bao2023channel, pham2024enhancing}, we train and evaluate CCE on ViTs \citep{dosovitskiy2020image}. Hierarchical attention-based encoders are being widely researched and show improvements over traditional ViTs \citep{fan2021multiscale, ryali2023hiera}. Therefore, we aim to adapt CCE to hierarchical architectures such as Hiera \citep{ryali2023hiera}, to build a general architectural framework beyond ViTs. Finally, we validate MCD using iBOT as our momentum-based SSL method \citep{zhou2021ibot}. We aim to evaluate the generalization of MCD to other momentum-based methods in the future. Although the use of MCD yields significant improvements on the ID tasks (by approximately 2\%), they do not translate to the OOD tasks. We aim to investigate this in future work.

\section{Conclusion}

In this paper, we introduce C3R, a two-fold framework comprising a branched encoder architecture and a novel training strategy for IHC images. C3R yields stronger cell image representations while enabling training- and adaptation-free out-of-distribution (OOD) evaluation. We show that C3R significantly outperforms in-distribution (ID) benchmarks and matches or exceeds baseline performance in OOD settings. By leveraging the context-concept principle, we improve upon existing channel-adaptive strategies and demonstrate the importance of encouraging concept channels to build strong representations with limited context. Furthermore, we validate our core assumption of context-concept separation through additional experiments and analysis. Overall, this work offers a new perspective on IHC datasets and opens a pathway for cross-dataset generalization without requiring dataset-specific adaptation or retraining.

\newpage

\input{appendix}

\newpage

{
	\small

\input{main.bbl}
}

\end{document}

%% file: appendix.tex
\appendix

\section*{Appendix}

We organize the supplementary material of the main paper as follows. \appref{sec:analysis} provides further insight into the context-concept assumption, design choices of C3R, and additional evaluations under channel-sparse settings. \appref{sec:datasets_implementation} describes in detail, the datasets and implementation protocols used for training and evaluation.

\section{Further analysis}
\label{sec:analysis}

\paragraph{Nomenclature. } We use the following definitions for the upcoming sections.
\[
\begin{array}{l l}
    x      & : \text{multi-channel input image}  \\
    h_{c1}, h_{c2} & : \text{context and concept convolutional stem} \\
    f_{c1}, f_{c2} & : \text{context and concept encoder} \\
    f_{s}   & : \text{shared encoder} \\
    \mathcal{S}, \mathcal{T} & : \text{student and teacher network}
\end{array}
\]

\subsection{The context-concept assumption in IHC images}
\label{sec:assumption}

To validate the assumption that the context channels are similar by texture across the instances, we compare the representations of each channel across a given dataset. 

Specifically, for a given multi-channel image, we first compute output representations for each channel through a ViT-B/14 pre-trained using DINOv2 on the LVD142M dataset \citep{oquab2023dinov2}. This is performed by obtaining a single image channel and repeating the image thrice along the channel dimension, to emulate an RGB image. We use a strong natural image checkpoint to prevent inducing a cell-image-specific bias when computing the representations, as our main focus is on the texture present in each channel.

\begin{figure}[ht]
    \centering
    \begin{subfigure}[b]{0.49\linewidth}
        \centering
        \includegraphics[width=\linewidth]{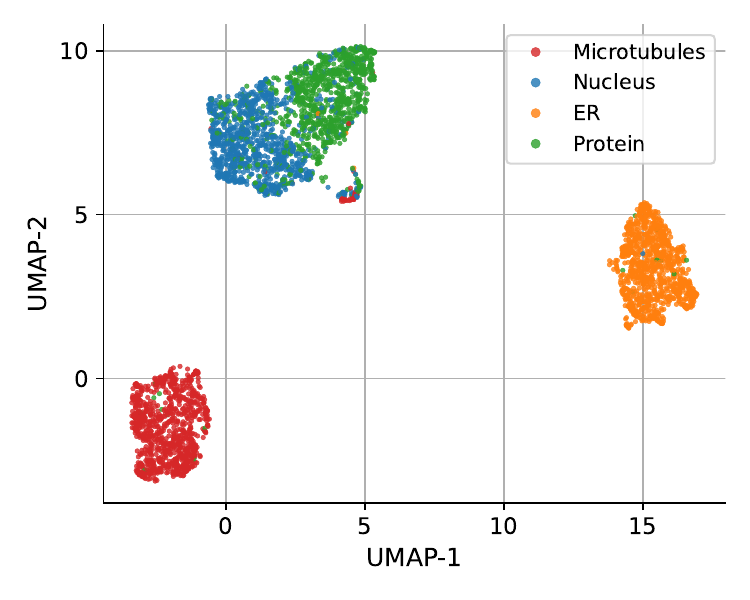}
        \caption{}
        \label{fig:umap}
        \end{subfigure}
    \begin{subfigure}[b]{0.49\linewidth}
        \centering
        \includegraphics[width=\linewidth]{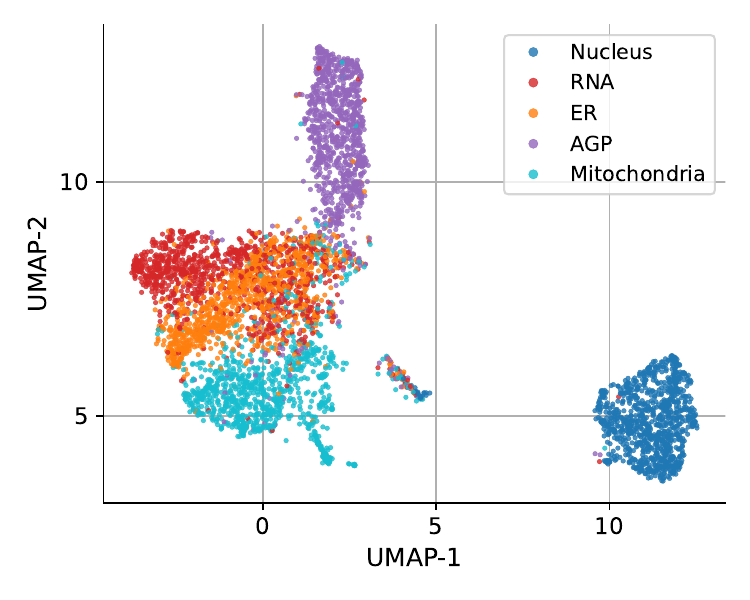}
        \caption{}
        \label{fig:kmeans}
        \end{subfigure}
    \caption{Visualization of channel-wise features in 2D space using UMAP on (a) HPA and (b) JUMP-CP. Most individual channels show clear inter-channel separation and intra-channel similarity between instances.}
    \label{fig:vis}
\end{figure}

We obtain such representations for \(N=1000\) random instances in the HPA and JUMP-CP datasets, which contain 4 and 5 channels, respectively. We then perform Universal Manifold Approximation and Projection (UMAP) \citep{mcinnes2018umap} on these features to visualize the representations in 2D, as seen in \fref{fig:vis}. Here, we observe that most individual channels show clear inter-channel separation and intra-channel
similarity across instances.

To quantify the observations from \fref{fig:vis}, we perform K-Means clustering \citep{lloyd1982least} on all features of all channels in a given dataset, using \(K=4\) clusters on HPA, and \(K=5\) clusters on JUMP-CP. Each channel is then assigned to the cluster in which its features appear most frequently. Then, the concentration of a channel in is assigned cluster, and its spread across other clusters is measured. 

Specifically, in a given dataset, for a given channel \(c\), we compute a discrete probability distribution of that channel's occurrence across the clusters as, 
\begin{align}
    \mathbf{p}_c = \{ p_c^k \mid k\in\{0,\dots,K-1\}\}, \quad \text{~with~} \quad p_c^k = \frac{1}{|\mathcal{I}_c|} \sum_{n \in \mathcal{I}_c} \mathbb{1}(c_n = k)
\end{align}
where \(\mathcal{I}_c = \{\,n \mid c_n = c\}\) is the set indices of samples that belong to channel \(c\). Here, \(p^c_k\) returns the fraction of samples of channel \(c\) that fall into cluster \(k\). We then compute the parity \(P_c\) and entropy \(H_c\) of each channel \(c\). In our case, the parity \(P_c\) indicates the concentration of a single channel \(c\) in its assigned cluster, whereas the entropy \(H_c\) indicates the spread of channel \(c\) across all clusters \citep{shannon1948mathematical}. \(H_c=0\) indicates a perfect match, while \(H_c=\log_2(K)\) where \(K\) is the cluster count, indicates a uniform distribution of a channel across all clusters.

We compute \(P_c\) and \(H_c\) as,
\begin{align}
    P_c =  \max \big(\mathbf{p}_c\big) \quad \text{~and~} \quad
    H_c = - \sum_{k=0}^{K-1} \big(  p_c^k \big) \cdot \log_2 \big( p_c^k  \big) \quad.
\end{align}

\tref{tab:parity} shows the computed parity \(P_c\) and entropy \(H_c\) for each channel for each dataset, where we observe a high parity for the context channels vs. the concept channels, and high entropy for the concept channels vs. the context channels. This validates our assumption that the context channels are statistically similar across instances in a dataset.  

\begin{table}[ht]
    \centering
    
    \captionof{table}{Parity \(P_c\) and entropy  \(H_c\)  of cluster-assigned channel features in HPA and JUMP-CP.}
    \label{tab:parity}

    \begin{tabular}{ll cccc}
    \toprule
    \multirow{2}{*}{Group} & \multirow{2}{*}{Channel} & \multicolumn{2}{c}{HPA} & \multicolumn{2}{c}{JUMP-CP} \\
    \cmidrule(lr){3-4} \cmidrule(lr){5-6}
    && \(P_c\) & \(H_c\) & \(P_c\) & \(H_c\) \\
    \midrule
    \multirow{3}{*}{Context}     &Microtubules & 0.942 & 0.374 & - & -\\
         &Nucleus      & 0.881 & 0.656 & 0.943 & 0.372\\
         &ER           & 0.983 & 0.140 & 0.489 & 1.496 \\
    \midrule
    \multirow{4}{*}{Concept}     &Protein      & 0.759 & 1.025 & - & -\\
         &RNA & - & -& 0.217 & 1.987 \\
         &AGP &- &-& 0.513 & 1.470 \\
         &Mitochondria & - & - & 0.357 & 1.918\\
    \bottomrule
    \end{tabular}

\end{table}

However, the parity of the ER channel in JUMP-CP is notably lower, and its entropy higher than expected for a context channel, although these metrics behave as expected in HPA. This is likely due to the overall entropy of JUMP-CP being higher than HPA, where JUMP-CP is relatively a noisy dataset compared to HPA which contains clear and high-resolution images. Such noise can blur the distinction between context and concept channels, leading to less clear clustering. We aim to investigate better methods of statistically identifying the context and concept channels in future work.

\subsection{Effect of the number of layers pre- and post-merging in the Context-Concept Encoder (CCE)}
\label{sec:design0}

The design of CCE leads to exploring the amount of separate per-group processing needed for the context and concept channels before merging. This dictates the layer depths of \(f_{c1}\) and \(f_{c2}\). Moreover, as we adjust the layer depth of \(f_s\) for CCE to maintain the total parameter count of a baseline ViT, the depth of \(f_s\) indirectly reflects the amount of joint processing carried out after merging the context and concept groups.

To investigate the extent of per-group processing needed, we first vary the number of layers of \(f_{c1}\) and \(f_{c2}\), and keep the number of layers of \(f_s\) adjusted to match the baseline parameter count. 

\begin{figure}[ht]
    \centering
    \begin{subfigure}[b]{0.42\linewidth}
        \centering
        \includegraphics[width=\linewidth]{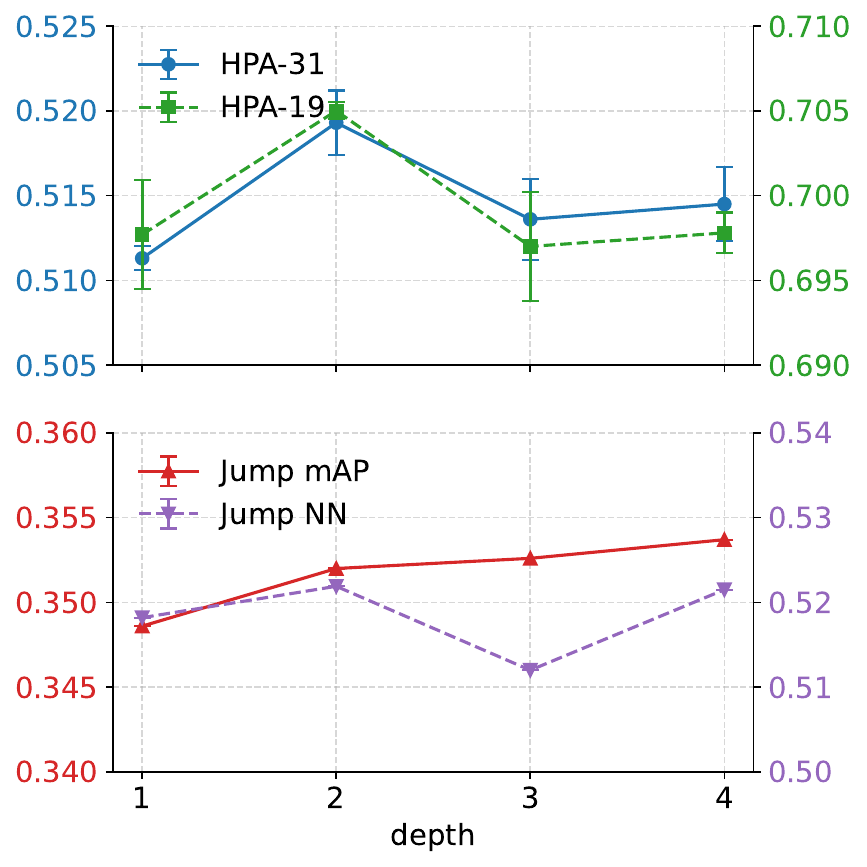}
        \caption{}
        \label{fig:pre_layers}
    \end{subfigure}
    \begin{subfigure}[b]{0.42\linewidth}
        \centering
        \includegraphics[width=\linewidth]{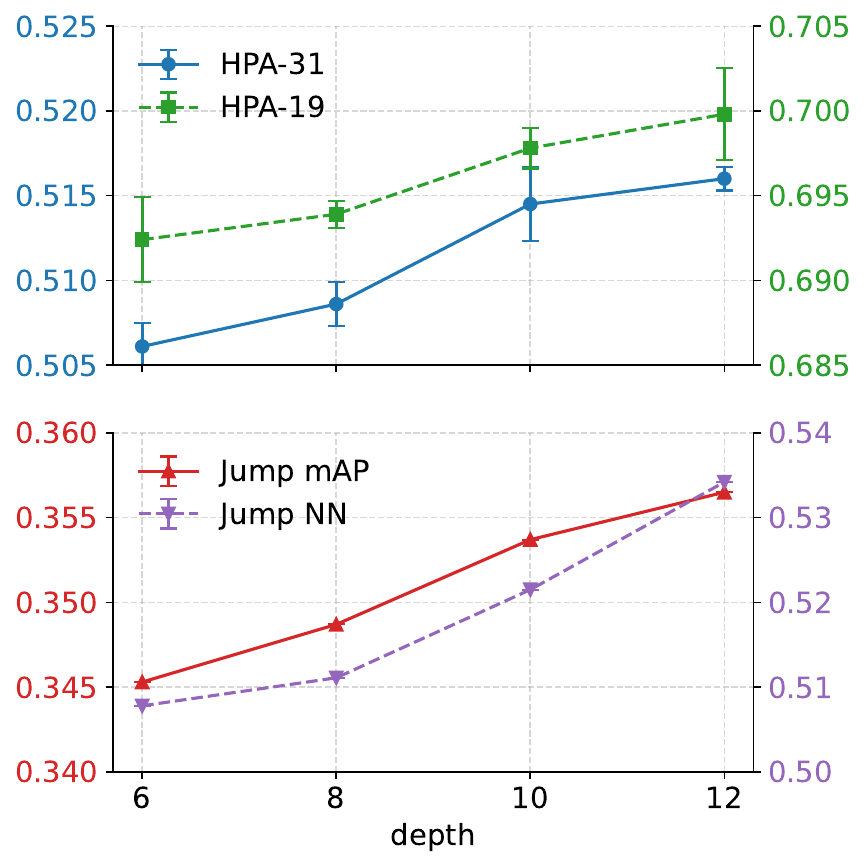}
        \caption{}
        \label{fig:post_layers}
    \end{subfigure}
    
    \caption{Comparison of performance impact from (a) pre-merging and (b) post-merging depths on HPA (top) and JUMP-CP (bottom). All experiments are performed on the CCE architecture with ViT-S and without Masked Context Distillation.}
    \label{fig:combined_layers}
\end{figure}

\fref{fig:pre_layers} shows that two layers each for \(f_{c1}\) and \(f_{c2}\) yields the best overall performance on HPA, whereas the effect of pre-merging depth on JUMP-CP is negligible. Given that the total parameter count is fixed, we investigate if the reduction in performance for HPA with higher depth for \(f_{c1}\) and \(f_{c2}\) is due to the loss of low-level feature granularity (caused by excessive abstraction before merging) or whether the adjusted number of combined layers of \(f_s\) is insufficient to build a strong global representation.

To explore this, we fix the number of \(f_{c1}\) and \(f_{c2}\) layers at 4 (the deepest branch depth experimented previously) and vary the depth of \(f_s\) from \([6, 8, 10, 12]\), with 12 being equivalent to the depth of a standard ViT backbone.

\fref{fig:post_layers} shows the effect of the depth of \(f_s\) given a fixed branch depth, where we observe that the performance increases with increasing depth for both HPA and JUMP-CP.  However, for HPA, the overall performance does not exceed that of a parameter-normalized network with two branched layers, implying that HPA benefits from preserving low-level features before merging. In contrast, JUMP-CP metrics show a global increase with increasing branch and shared depth, indicating that JUMP-CP benefits from both a deeper network due to more parameters, and a more abstract representation of groups
with the use of deeper branches.

We suspect that this is mainly due to the difference in granularity between in datasets: HPA is inherently more fine-grained than JUMP-CP, hence HPA benefits from low level information even after merging, while JUMP-CP does not.

\subsection{Pre vs. Post-aggregation and its effects on Masked Context Distillation (MCD)}
\label{sec:design1}

\begin{figure}[t]
    \centering
    \begin{subfigure}[]{0.45\linewidth}
        \centering
        \includegraphics[width=\linewidth]{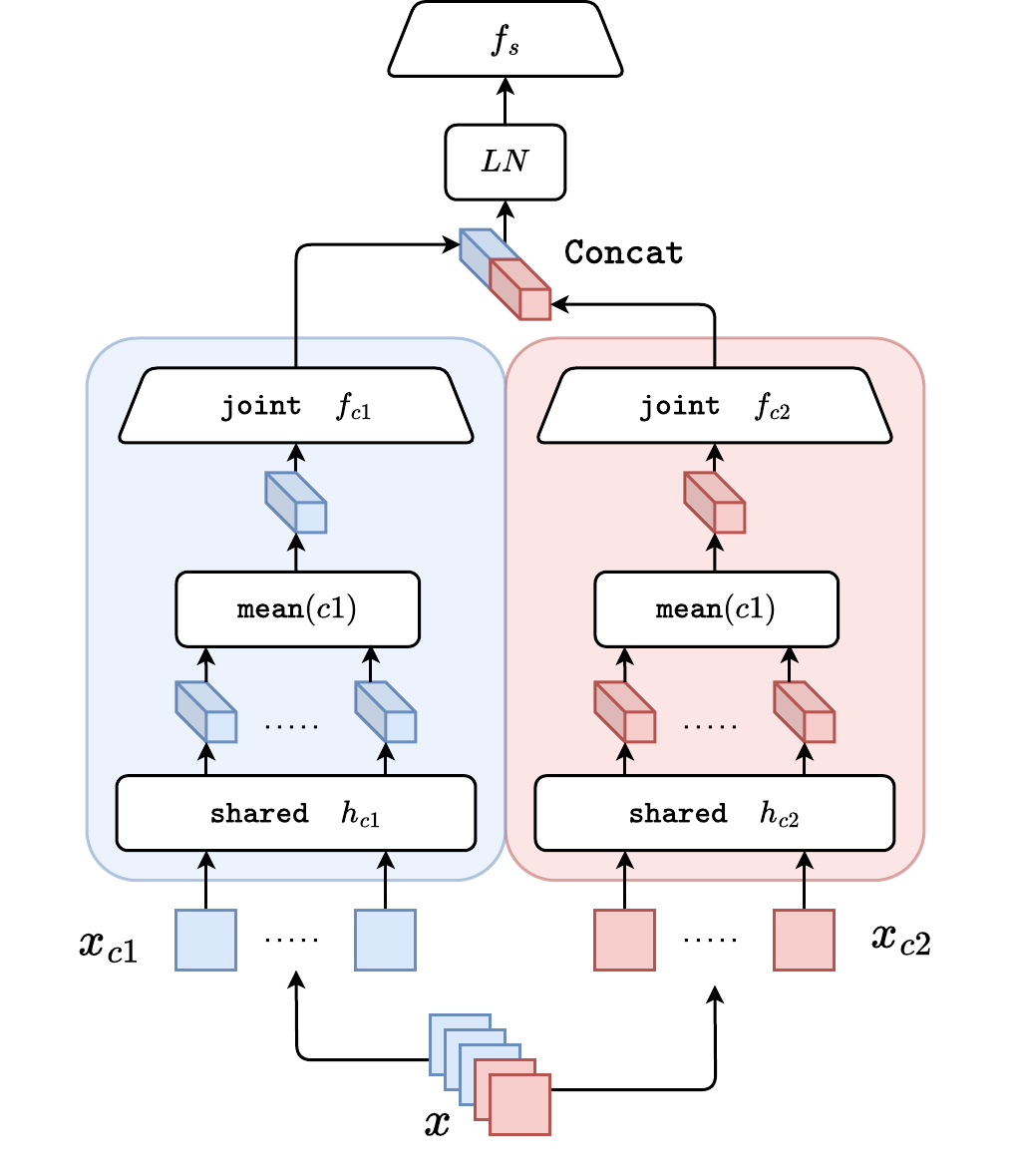}
        \caption{}
        \label{fig:pre_agg}
    \end{subfigure}
    \begin{subfigure}[]{0.45\linewidth}
        \centering
        \includegraphics[width=\linewidth]{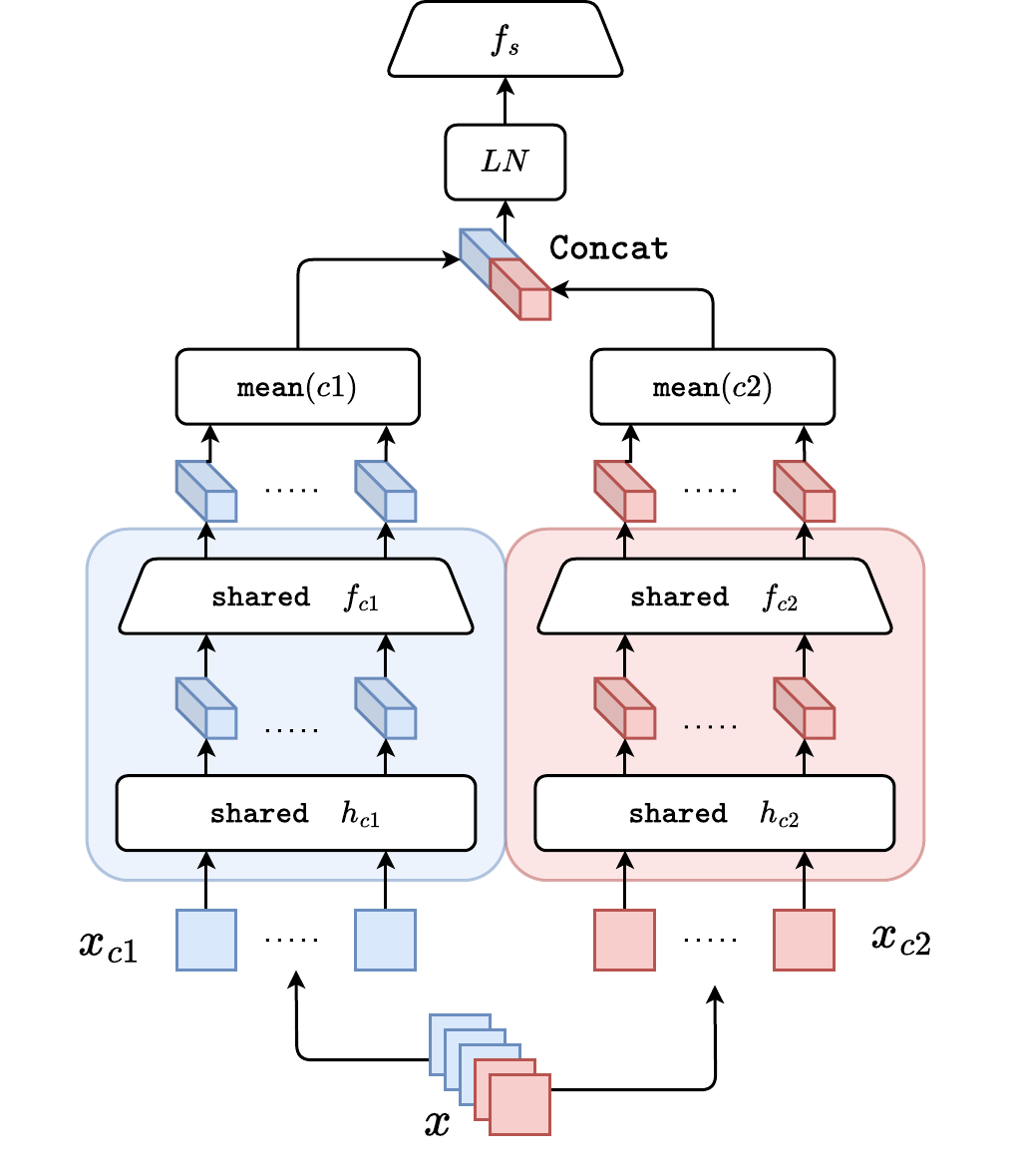}
        \caption{}
        \label{fig:post_agg}
    \end{subfigure}
    
    \caption{(a) Pre-aggregation. The channels are passed separately through the convolutional stem, mean-pooled and then jointly passed through the encoder branch. (b) Post-aggregation. The channels are passed separately through the convolutional stem and the encoder branch, then mean-pooled.}
    \label{fig:pre_post}
\end{figure}

We investigate on how the aggregation method (pre or post-aggregation) of the CCE architecture effects overall performance, particularly when MCD is applied. \fref{fig:pre_agg} and \fref{fig:post_agg} show in detail, the architecture of pre and post-aggregation variants, respectively. 
\begin{table}[b]
        \centering
        \captionof{table}{Effects of pre- vs. post-aggregation under MCD and no MCD. All experiments are performed on ViT-S.}
        \label{tab:mcd-prepost}
        \begin{tabular}{l l cccc}
        \toprule
        \multirow{2}{*}{MCD} & \multirow{2}{*}{Aggr.} & \multicolumn{2}{c}{HPA-mAP} & \multicolumn{2}{c}{JUMP-CP} \\
        \cmidrule(lr){3-4} \cmidrule(lr){5-6}
        & & 31-loc & 19-loc & mAP & kNN \\
        \midrule
        \multicolumn{2}{@{}l}{\gray{Baseline}} & \gray{0.505} & \gray{0.686} & \gray{\underline{0.355}} & \gray{\underline{0.507}} \\
        \midrule
        \multirow{2}{*}{\xmark} 
            & Pre     & \textbf{0.520} & \textbf{0.705} & \textbf{0.358} & \textbf{0.530} \\
            & Post    & 0.516 & 0.702 & 0.351 & 0.529 \\
        \midrule
        \multirow{2}{*}{\cmark} 
            & Pre     & 0.527 & 0.713 & 0.350 & 0.508 \\
            & Post    & \textbf{0.536} & \textbf{0.722} & \textbf{0.354} & \textbf{0.518} \\
        \bottomrule
        \end{tabular}%
\end{table}

\tref{tab:mcd-prepost} shows that pre-aggregation performs better without MCD. In pre-aggregation, the branched encoders \(f_{c1}\) and \(f_{c2}\) learn from a joint representation of all the channels that belong to the group. Hence, unlike post-aggregation, \(f_{c1}\) and \(f_{c2}\) learn intra-group relationships, which result in higher ID and OOD metrics. Therefore, we use the pre-aggregation variant whenever MCD is not used.

In contrast, with MCD, the post-aggregation variant outperforms the pre-aggregation variant. We hypothesize that the MCD loss benefits from diverse student-teacher representations, and the independent modeling of each channel through post-aggregation yields these diverse representations. 

To validate this hypothesis, we study the output representations for full vs. sparse context inputs, for both pre-aggregation and post-aggregation variants. Specifically, for a given variant, we compare the \texttt{cls} token representations from full vs. sparse inputs at two stages—the intermediate output after the context encoder and the final output after the shared encoder. 

First, we view the full and sparse inputs \(x\) and \(\bar{x}\) as, 
\begin{align}
x_{c1-\texttt{drop}} &= \texttt{DropChannels}(x_{c1}, c)  \in \mathbb{R}^{B \times (C_1 - c) \times h \times w}, \quad \text{~where~} 0 \leq c < C_1 \\
x &= [x_{c1},\, x_{c2}] , \quad  \quad \bar{x} = [x_{c1-\texttt{drop}},\, x_{c2}] .
\end{align}

We pass the inputs \(x\) and \(\bar{x}\) through the convolutional stems and branched encoders of an MCD pre-trained teacher \(\mathcal{T}\), to obtain the full and sparse intermediate context representations \(p_{c1}\) and \(\bar{p}_{c1}\) and the full and sparse final output representations \(y\) and \(\bar{y} \). As we do not perform channel dropping for the concept group, we do not consider the intermediate representations from \(h_{c2}\) and \(f_{c2}\) for our analysis.
\begin{align}
     p_{c1} &=  f_{c1}\big(h_{c1}(x_{c1})\big) \quad  &\bar{p}_{c1} & =  f_{c1}\big(h_{c1}(x_{c1-\mathtt{drop}})\big)\\
      y &=  \mathcal{T}\big(x \big) \quad &\bar{y} &=  \mathcal{T}\big(\bar{x} \big) \quad 
\end{align}

\begin{figure}[ht]
    \centering
    \begin{subfigure}[b]{0.45\linewidth}
        \centering
        \includegraphics[width=\linewidth]{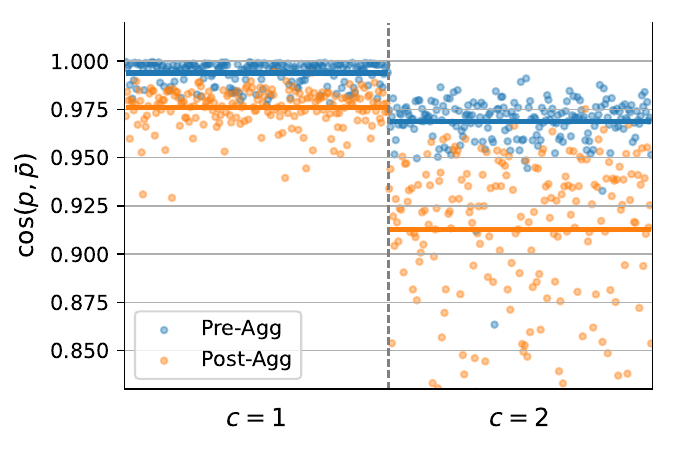}
        \caption{}
        \label{fig:cos_inter}
    \end{subfigure}
    \begin{subfigure}[b]{0.45\linewidth}
        \centering
        \includegraphics[width=\linewidth]{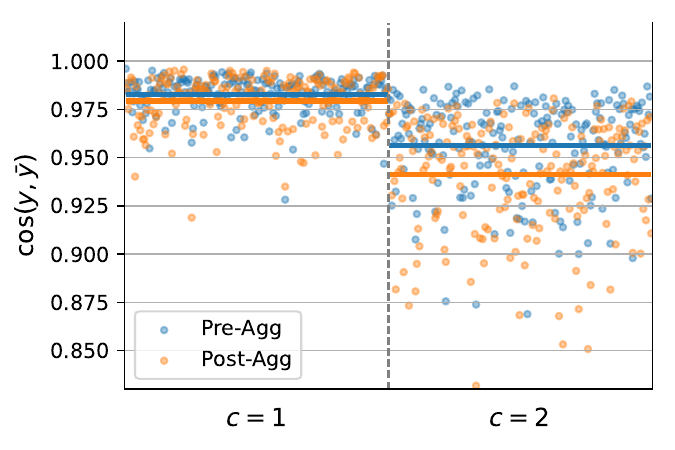}
        \caption{}
        \label{fig:cos_final}
    \end{subfigure}
    
    \caption{(a) Cosine similarity between full and sparse intermediate representations \(p\) and \(\bar{p}\). (b) Cosine similarity between full and sparse output representations \(y\) and \(\bar{y}\).}
    \label{fig:cos_sim}
\end{figure}

We then compute the cosine similarity between \(p_{c1}\) and \( \bar{p}_{c1}\), and the cosine similarity between \(y\) and \( \bar{y}\). \fref{fig:cos_sim} shows the cosine similarities for both intermediate and final outputs, computed on \(N=200\) random instances from the HPA dataset, under both the pre-aggregation and post-aggregation variants. We plot the cases where \(c=1\) and \(c=2\) channels are dropped to create the sparse input \(\bar{x}\).

Both \fref{fig:cos_inter} and \fref{fig:cos_final} show that for both intermediate and final outputs, similarity reduces with increased sparsity i.e: higher channel dropping. However, from \fref{fig:cos_inter}, we observe that the pre-aggregation intermediate outputs show much higher similarity, implying that \(f_{c1}\) returns a very similar representation with and without dropping. Therefore, the concept channels \(x_{c2}\) of the student and teacher attend to similar context representations during the training step, yielding similar final representations without additional work done by the shared encoder \(f_s\).

In contrast, \fref{fig:cos_inter} shows that post aggregation of channels yields more diverse intermediate representations than pre-aggregation, indicating that the sparse context input translates to a sparse context representation. This forces the concept channels \(x_{c2}\) to contribute to the same global representation from dissimilar contexts (full vs. sparse) during the training step, correctly following the motivation of MCD.

\subsection{Context channel sampling rates during MCD training}
\label{sec:design2}

Using the post-aggregation variant, we explore the optimal sampling rate for the context channels when training with MCD. Specifically, we run fixed sampling rates where we drop \(c=1\) and \(c=2\) channels, and a variable sampling rate where we randomly drop \( c \in \{0, 1, 2\} \) context channels, sampled uniformly at random. The total number of context channels in the HPA dataset is 3 (Microtubules, ER and Nucleus). 
\begin{table}[ht]
        \centering
        \captionof{table}{Effect of context channel sampling rates in during MCD training. All experiments are performed on ViT-S using the post-aggregation variant.}
        \label{tab:mcd-sparsity}
        \begin{tabular}{l c c c c}
        \toprule
        \multirow{2}{*}{\(c\)} & \multicolumn{2}{c}{HPA-mAP} & \multicolumn{2}{c}{JUMP-CP} \\
        \cmidrule(lr){2-3} \cmidrule(lr){4-5}
         & 31-loc & 19-loc & mAP & kNN \\
        \midrule
        \gray{Baseline}                     & \gray{0.505} & \gray{0.686}  &   \gray{\underline{0.355}}  & \gray{\underline{0.507}}  \\
        \midrule
        $0$     &   0.516 &  0.702    &   0.351    &    \textbf{0.529}    \\
        $1$     & 0.531     &   0.708     &  \textbf{0.356}    & 0.521       \\
        $2$     & 0.513     &   0.700     &   0.336    & 0.499       \\
        \(\{0,1,2\}\)& \textbf{0.536}     &   \textbf{0.722}     &   0.354    & 0.518       \\
        \bottomrule
        \end{tabular}%
\end{table}

\tref{tab:mcd-sparsity} shows the effect of context channel sampling rates on ID and OOD performance. Here, we find that a random sampling rate \(c=\{0,1,2\}\) yields better overall performance vs. fixed sampling. Consistently sparse sampling at \(c=2\) exhibits the lowest performance, possibly because the distillation task becomes too difficult. In contrast, dropping only a single channel at \(c=1\) yields similar OOD metrics but lower ID metrics. This result, along with the higher metrics obtained with variable sampling rates, suggests that the student \(\mathcal{S}\) can still benefit from sparse context \(c = 2\), provided it receives enough support from other training iterations where fewer channels are dropped (e.g., \(c = 0\) or \(c = 1\)).

\subsection{Limiting observed channels during evaluation}
\label{sec:limit_eval}

 Here, we study the translation of MCD learning from limited context during training, to inferring from limited context during evaluation. The HPA and JUMP-CP datasets contain three and two context channels, respectively. We perform evaluation on both datasets by dropping out a context channel, forcing the concept channels to contribute to the representation with limited context.

\tref{tab:mcd_inference} shows the the effect of limited context channels during evaluation. We find that for HPA, when no MCD is applied, we there is a significant drop in performance. In contrast, MCD is robust to limited context, yielding very similar metrics with and without channel dropping. 

However, MCD's robustness to channel dropping does not translate to OOD evaluation, as we observe a consistent drop in metrics with and without MCD, on JUMP-CP. Although we suspect that this is due to the noise in JUMP-CP causing the context representations with and without channel dropping to collapse into one representation, this warrants further investigation, which we aim to carry out in future work.  
\begin{table}[ht]
        \centering
        \captionof{table}{Effect of limited context channels during evaluation. -[chan] corresponds to a given channel being dropped during evaluation. Arrows indicate average change relative to the ``All'' row for each method. As mentioned previously, for the encoder, we use pre-aggregation without MCD and post-aggregation with MCD. }
        \label{tab:mcd_inference}
        \begin{tabular}{l l llll}
        \toprule
        \multirow{2}{*}{MCD} & \multirow{2}{*}{Channels} & \multicolumn{2}{c}{HPA-mAP} & \multicolumn{2}{c}{JUMP-CP} \\
        \cmidrule(lr){3-4} \cmidrule(lr){5-6}
         & & 31-loc & 19-loc & mAP & kNN \\
        \midrule
         \multicolumn{2}{l}{\gray{Baseline}}                     & \gray{0.505} & \gray{0.686}  &   \gray{\underline{0.355}}  & \gray{\underline{0.507}}  \\
        \midrule
        \multirow{5}{*}{\xmark}    & All          &  0.520  & 0.705     &   0.358    &   0.530     \\
                                & \(-\)[Nuc]   &  0.515  & 0.697  & 0.352  & 0.529  \\
                                & \(-\)[ER]    &  0.481  & 0.650  & 0.334  & 0.509 \\
                                & \(-\)[MT]    & 0.494 & 0.679 & -- & -- \\
                                \cmidrule(lr){2-6}
& \(-\)[Average] & 0.497 \red{↓{\scriptsize \textbf{0.023}}} & 0.675 \red{↓{\scriptsize \textbf{0.030}}} & \textbf{0.343} \red{↓{\scriptsize 0.015}} & \textbf{0.519} \red{↓{\scriptsize 0.011}} \\

        \midrule
        \multirow{5}{*}{\cmark}    & All          & 0.536  & 0.722     & 0.354     & 0.518     \\
                                & \(-\)[Nuc]   & 0.543  & 0.723  & 0.336  & 0.509 \\
                                & \(-\)[ER]    & 0.527  & 0.713  & 0.339  & 0.501  \\
                                & \(-\)[MT]    & 0.524 & 0.709 & -- & -- \\
                                \cmidrule(lr){2-6}
& \(-\)[Average] & \textbf{0.531} \red{↓{\scriptsize 0.005}} & \textbf{0.715} \red{↓{\scriptsize 0.007}} & 0.338 \red{↓{\scriptsize \textbf{0.016}}} & 0.505 \red{↓{\scriptsize \textbf{0.013}}} \\

        \bottomrule
        \end{tabular}
\end{table}
\newpage

\section{Datasets and implementation}
\label{sec:datasets_implementation}

\subsection{Datasets}

\paragraph{Human Protein Atlas (HPA). } 

We use the Human Protein Atlas (HPA) Subcellular Localization dataset (version 23) \citep{thul2017subcellular}, a large-scale collection of immunofluorescence microscopy images capturing the expression and spatial distribution of 13,141 genes across 37 human cell lines. Each sample is stained with fluorescent markers targeting the nucleus, microtubules, endoplasmic reticulum, and a protein of interest, resulting in four-channel images that capture distinct structural and functional features of the cell. Protein localization is annotated across 35 subcellular compartments, covering both major organelles and fine-grained structures.

We extract single-cell instances using segmentation masks generated by the recommended HPA-Cell-Segmentation pipeline \citep{gupta2024subcell}. This yields a total of 1,138,378 single-cell crops. Data are split into training, validation, and test sets based on antibody identity to prevent information leakage across sets.

\paragraph{JUMP-CP. }

In this study, we use a subset of the JUMP-Cell Painting (JUMP1) pilot dataset \citep{chandrasekaran2024three}. The full dataset includes approximately 300 million images of U2OS and A549 cells subjected to diverse chemical and genetic perturbations. Following \citep{gupta2024subcell}, we focus on chemical perturbations applied to the U2OS cell line. The Cell Painting assay stains eight cellular components using six fluorescent dyes, imaged across five channels: nucleus, mitochondria, nucleoli and cytoplasmic RNA, endoplasmic reticulum, Golgi and plasma membrane, and actin cytoskeleton.

We select seven plates for analysis: BR00117010, BR00117011, BR00110712, BR00117013, BR00117024, BR00117025, and BR00110726. Each plate follows an identical treatment layout, consisting of 64 wells with negative controls and 320 wells treated with one of 302 unique compounds. Images from these plates are preprocessed and segmented into single-cell crops using DeepProfiler with default settings, following Subcell
\citep{gupta2024subcell}.

Both HPA and JUMP-CP data are publicly available online\footnote{\url{https://www.proteinatlas.org}}\footnote{\url{https://github.com/broadinstitute/cellpainting-gallery}}.

\subsection{Implementation details}

The C3R training pipeline can be found at: \url{https://anonymous.4open.science/r/C3R-5015}.

All single-cell crops are resized to \(224 \times 224\) for HPA pre-training, \(448 \times 448\) for HPA linear evaluation, and \(112 \times 122\) for JUMP-CP. Inputs are [0, 1] normalized per-channel, unless otherwise stated.

\paragraph{Pre-training on HPA.} 
We pre-train all models using a Vision Transformer (ViT) backbone \citep{dosovitskiy2020image}, following the iBOT framework \citep{zhou2021ibot}\footnote{\url{https://github.com/bytedance/ibot}}, on the HPA dataset \citep{thul2017subcellular}. The dataset comprises \(\mathrm{N_I} = 1138378\) single-cell images annotated with \(\mathrm{N_A}=11920\) antibodies. All cells with a unique antibody ID are distributed across 4  samples in the dataset. This results in an average of \(\mathrm{N_I} / (4 \times \mathrm{N_A}) \approx 24\) cells per sample, and \(4 \times \mathrm{N_A} = 47680\) overall samples.

Following \citet{gupta2024subcell}, we incorporate their antibody supervision loss, added directly to the iBOT objective without reweighting. Each iteration samples 8 cells per training sample, producing 8 positives and \(B{-}8\) negatives for the supervised contrastive loss (where \(B\) is the total batch size). The antibody loss and dataset utilities are obtained from the official SubCell repository\footnote{\url{https://github.com/CellProfiling}}.

For ViT-S/16 and ViT-B/16 encoders, we use AdamW optimizer \citep{loshchilov2017decoupled} with a base learning rate of \(5 \times 10^{-4}\), cosine annealing, and 10 warm-up epochs. ViT-S is trained for 100 epochs using a patch masking ratio sampled from \([0.1, 0.5]\) (mean \(0.3 \pm 0.2\)); ViT-B is trained for 50 epochs with masking ratio \([0.1, 0.7]\) (mean \(0.4 \pm 0.3\)). Each image undergoes 2 global crops and 10 local crops. We use batch sizes of 80 (ViT-S) and 48 (ViT-B) per GPU, training on 8 RTX 3090 GPUs.

For ChannelViT and DiChaViT \citep{bao2023channel, pham2024enhancing}, we adapt their official model and loss implementations\footnote{\url{https://github.com/insitro/ChannelViT}} \footnote{\url{https://github.com/chaudatascience/diverse_channel_vit}} to the iBOT pipeline. For both methods, we sample 2 channels per iteration. For DiChaViT, the first channel is randomly selected and shared between student and teacher networks.

For CCE and C3R (CCE + MCD), we define 2 branched layers per group at half the embedding dimension, leading to 11 shared layers with full embedding dimension under parameter normalization. In MCD, context channels are dropped from both global and local crop inputs to the student branch. All other settings match the base iBOT configuration.

\paragraph{Linear evaluation on HPA.} We perform 19-class and 31-class multi-label classification on the field-of-view (FoV) level on the HPA dataset. A 3-layer MLP is trained for all evaluations. We use the official Subcell evaluation code \citep{gupta2024subcell}, which aggregates single-cell predictions to the field-of-view (FoV) level. Single-cell input size of 448 by 448 pixels was used during feature extraction prior to training the MLP. Models are trained using the Adam optimizer \citep{kingma2014adam} with a learning rate of \(1e^{-3}\), Sigmoid Focal Loss \citep{lin2017focal}, and early stopping based on validation mAP. Training is run for up to 100 epochs.

\paragraph{Zero-shot evaluation on JUMP-CP.} For zero-shot evaluation, we extract embeddings from each single-cell image at a resolution of \(112 \times 122\) pixels. The resulting embedding has dimensionality \(3 \times d\) where \(d\) is the model embedding dimension, as the Mitochondria, RNA, and AGP channels are each processed independently in combination with the Nucleus and ER channels, and their outputs are concatenated. The concatenated embeddings are then subjected to dimensionality reduction.

To generate well-level profiles, we follow a two-stage aggregation strategy: first, cell-level embeddings are averaged to form Field-of-View (FoV) representations; second, FoV embeddings are averaged to produce well-level profiles. For post-processing, we follow the SubCell protocol \citep{gupta2024subcell}, selecting the best combination from a set of dimensionality reduction methods \{\textsc{PCA}, \textsc{ZCA}\} \citep{kessy2018optimal} and normalization techniques \{MAD-based robust standardization, z-score normalization, None\}. We report results using the configuration that yields the highest performance. We report the k-NN evaluation and mAP metrics on replicate retrieval, where a positive replicate is subject to the same perturbation (out of 302 perturbations) as the query.

\paragraph{CHAMMI benchmark. }
We use the official CHAMMI repository\footnote{\url{https://github.com/chaudatascience/channel_adaptive_models}} and follow the benchmarking setup described in the original paper \citep{chen2023chammi}. All models are fine-tuned using a ConvNeXt backbone \citep{liu2022convnet} pre-trained on ImageNet-22k. For training the proposed context-concept stem network, we use the same hyper-parameters recommended for HyperNet. When the model views WTC-11 images, the Nucleus channel is set as context, and Membrane and Protein channels as concept.